\documentclass[10pt,twocolumn,letterpaper]{article}

\usepackage{cvpr,times,graphicx,authblk,amsmath,amssymb,tabulary,multirow}
\usepackage[pagebackref=false,breaklinks=true,letterpaper=true,colorlinks=true,bookmarks=false]{hyperref}
\usepackage[british,UKenglish,USenglish,english,american]{babel}

\newcommand{\bd}[1]{\textbf{#1}}

\def\x{$\times$}
\newlength\savewidth\newcommand\shline{\noalign{\global\savewidth\arrayrulewidth
  \global\arrayrulewidth 1pt}\hline\noalign{\global\arrayrulewidth\savewidth}}

\makeatletter
\renewcommand\paragraph{\@startsection{paragraph}{4}{\z@}%
  {.5em \@plus1ex \@minus.2ex}%
  {-1em}%
  {\normalfont\normalsize\bfseries}}
\makeatother

\cvprfinalcopy
\begin{document}

\title{Feature Pyramid Networks for Object Detection\vspace{-2mm}}

\author[1,2]{Tsung-Yi Lin}
\author[1]{Piotr  Doll\'ar}
\author[1]{Ross Girshick}
\author[1]{\\Kaiming He}
\author[1]{Bharath Hariharan}
\author[2]{Serge Belongie}
\affil[1]{Facebook AI Research (FAIR)}
\affil[2]{Cornell University and Cornell Tech\vspace{-1mm}}

\maketitle

\begin{abstract}
Feature pyramids are a basic component in recognition systems for detecting objects at different scales.
But recent deep learning object detectors have avoided pyramid representations, in part because they are compute and memory intensive.
In this paper, we exploit the inherent multi-scale, pyramidal hierarchy of deep convolutional networks to construct feature pyramids with marginal extra cost.
A top-down architecture with lateral connections is developed for building high-level semantic feature maps at all scales.
This architecture, called a Feature Pyramid Network (FPN), shows significant improvement as a generic feature extractor in several applications. 
Using FPN in a basic Faster R-CNN system, our method achieves state-of-the-art single-model results on the COCO detection benchmark without bells and whistles, surpassing all existing single-model entries including those from the COCO 2016 challenge winners.
In addition, our method can run at 6 FPS on a GPU and thus is a practical and accurate solution to multi-scale object detection. Code will be made publicly available.
\end{abstract}

\section{Introduction}

\begin{figure}[t]
\centering
\includegraphics[width=1.0\linewidth]{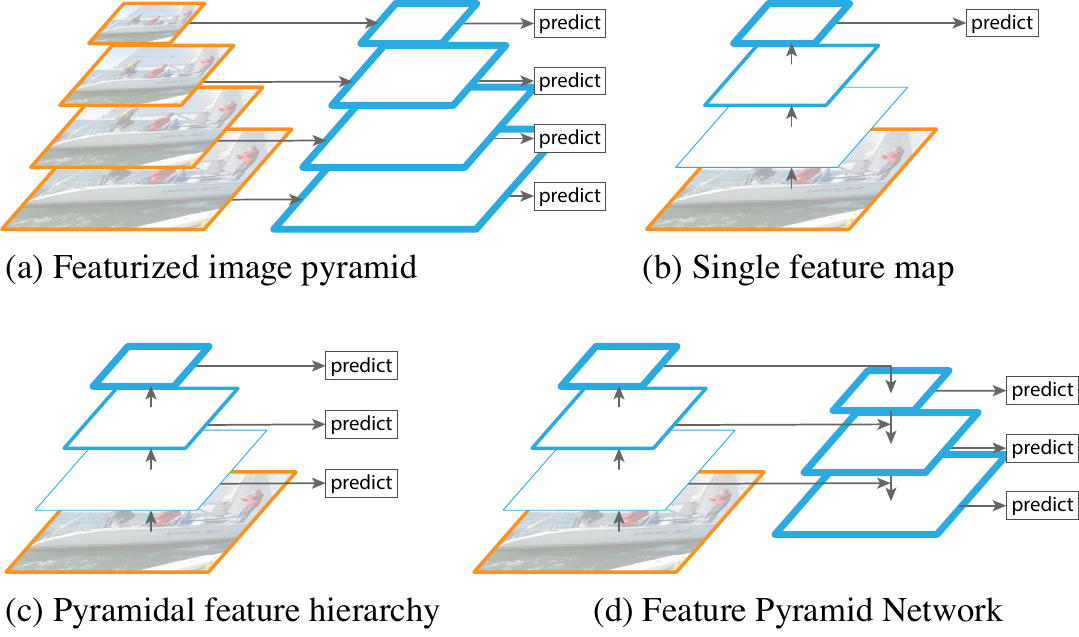}
\caption{(a) Using an image pyramid to build a feature pyramid. Features are computed on each of the image scales independently, which is slow. (b) Recent detection systems have opted to use only single scale features for faster detection. (c) An alternative is to reuse the pyramidal feature hierarchy computed by a ConvNet as if it were a featurized image pyramid. (d) Our proposed Feature Pyramid Network (FPN) is fast like (b) and (c), but more accurate. In this figure, feature maps are indicate by blue outlines and thicker outlines denote semantically stronger features.}
\label{fig:teaser}
\end{figure}

Recognizing objects at vastly different scales is a fundamental challenge in computer vision.
\emph{Feature pyramids built upon image pyramids} (for short we call these \emph{featurized image pyramids}) form the basis of a standard solution \cite{Adelson1984} (Fig.~\ref{fig:teaser}(a)).
These pyramids are scale-invariant in the sense that an object's scale change is offset by shifting its level in the pyramid.
Intuitively, this property enables a model to detect objects across a large range of scales by scanning the model over both positions and pyramid levels.

Featurized image pyramids were heavily used in the era of hand-engineered features \cite{Dalal2005,Lowe2004}.
They were so critical that object detectors like DPM \cite{Felzenszwalb2010} required dense scale sampling to achieve good results (\eg, 10 scales per octave).
For recognition tasks, engineered features have largely been replaced with features computed by deep convolutional networks (ConvNets) \cite{Krizhevsky2012,LeCun1989}.
Aside from being capable of representing higher-level semantics, ConvNets are also more robust to variance in scale and thus facilitate recognition from features computed on a single input scale \cite{He2014,Girshick2015a,Ren2015a} (Fig.~\ref{fig:teaser}(b)).
But even with this robustness, pyramids are still needed to get the most accurate results.
All recent top entries in the ImageNet \cite{Russakovsky2015} and COCO \cite{Lin2014} detection challenges use multi-scale testing on featurized image pyramids (\eg, \cite{He2016,Shrivastava2016}).
The principle advantage of featurizing each level of an image pyramid is that it produces a multi-scale feature representation in which \emph{all levels are semantically strong}, including the high-resolution levels.

Nevertheless, featurizing each level of an image pyramid has obvious limitations.
Inference time increases considerably (\eg, by four times \cite{Girshick2015a}), making this approach impractical for real applications.
Moreover, training deep networks end-to-end on an image pyramid is infeasible in terms of memory, and so, if exploited, image pyramids are used only at test time \cite{He2014,Girshick2015a,He2016,Shrivastava2016}, which creates an inconsistency between train/test-time inference.
For these reasons, Fast and Faster R-CNN \cite{Girshick2015a,Ren2015a} opt to not use featurized image pyramids under default settings.

However, image pyramids are not the only way to compute a multi-scale feature representation.
A deep ConvNet computes a \emph{feature hierarchy} layer by layer, and with subsampling layers the feature hierarchy has an inherent multi-scale, pyramidal shape.
This in-network feature hierarchy produces feature maps of different spatial resolutions, but introduces large semantic gaps caused by different depths.
The high-resolution maps have low-level features that harm their representational capacity for object recognition.

The Single Shot Detector (SSD) \cite{Liu2016} is one of the first attempts at using a ConvNet's pyramidal feature hierarchy as if it were a featurized image pyramid (Fig.~\ref{fig:teaser}(c)). Ideally, the SSD-style pyramid would reuse the multi-scale feature maps from different layers computed in the forward pass and thus come free of cost.
But to avoid using low-level features SSD foregoes reusing already computed layers and instead builds the pyramid starting from high up in the network (\eg,  conv4\_3 of VGG nets \cite{Simonyan2015}) and then by adding several new layers.
Thus it misses the opportunity to reuse the higher-resolution maps of the feature hierarchy.
We show that these are important for detecting small objects.

The goal of this paper is to naturally leverage the pyramidal shape of a ConvNet's feature hierarchy while creating a feature pyramid that has strong semantics at all scales.
To achieve this goal, we rely on an architecture that combines low-resolution, semantically strong features with high-resolution, semantically weak features via a top-down pathway and lateral connections (Fig.~\ref{fig:teaser}(d)).
The result is a feature pyramid that has rich semantics at all levels and is built quickly from a single input image scale.
In other words, we show how to create in-network feature pyramids that can be used to replace featurized image pyramids without sacrificing representational power, speed, or memory.

\begin{figure}[t]
\centering
\includegraphics[width=0.74\linewidth]{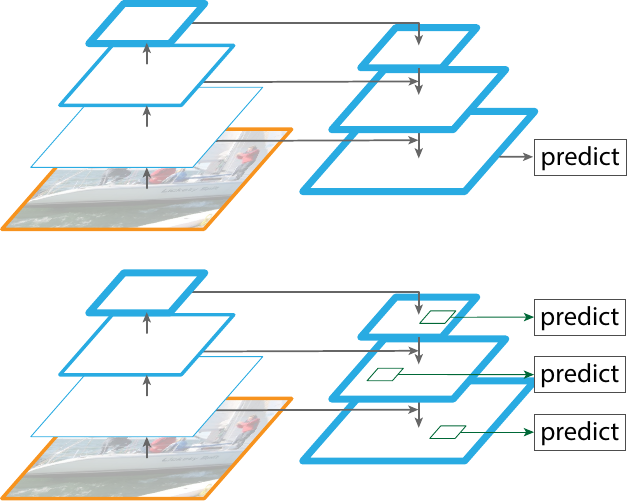}
\caption{Top: a top-down architecture with skip connections, where predictions are made on the finest level (\eg, \cite{Pinheiro2016}). Bottom: our model that has a similar structure but leverages it as a \emph{feature pyramid}, with predictions made independently at all levels.}
\label{fig:multilevel}
\end{figure}

Similar architectures adopting top-down and skip connections are popular in recent research \cite{Pinheiro2016,Honari2016,Ghiasi2016,Newell2016}.
Their goals are to produce a single high-level feature map of a fine resolution on which the predictions are to be made (Fig.~\ref{fig:multilevel} top).
On the contrary, our method leverages the architecture as a feature pyramid where predictions (\eg, object detections) are independently made on each level (Fig.~\ref{fig:multilevel} bottom). Our model echoes a featurized image pyramid, which has not been explored in these works.

We evaluate our method, called a Feature Pyramid Network (FPN), in various systems for detection and segmentation \cite{Girshick2015a,Ren2015a,Pinheiro2015}.
Without bells and whistles, we report a state-of-the-art single-model result on the challenging COCO detection benchmark \cite{Lin2014} simply based on FPN and a basic Faster R-CNN detector \cite{Ren2015a}, surpassing all existing heavily-engineered single-model entries of competition winners. 
In ablation experiments, we find that for bounding box proposals, FPN significantly increases the Average Recall (AR) by 8.0 points; for object detection, it improves the COCO-style Average Precision (AP) by 2.3 points and PASCAL-style AP by 3.8 points, over a strong single-scale baseline of Faster R-CNN on ResNets \cite{He2016}.
Our method is also easily extended to mask proposals and improves both instance segmentation AR and speed over state-of-the-art methods that heavily depend on image pyramids.

In addition, our pyramid structure can be trained end-to-end with all scales and is used consistently at train/test time, which would be memory-infeasible using image pyramids.
As a result, FPNs are able to achieve higher accuracy than all existing state-of-the-art methods.
Moreover, this improvement is achieved without increasing testing time over the single-scale baseline.
We believe these advances will facilitate future research and applications.
Our code will be made publicly available.

\section{Related Work}

\paragraph{Hand-engineered features and early neural networks.}
SIFT features \cite{Lowe2004} were originally extracted at scale-space extrema and used for feature point matching.
HOG features \cite{Dalal2005}, and later SIFT features as well, were computed densely over entire image pyramids.
These HOG and SIFT pyramids have been used in numerous works for image classification, object detection, human pose estimation, and more.
There has also been significant interest in computing featurized image pyramids quickly.
Doll{\'a}r \etal \cite{Dollar2014} demonstrated fast pyramid computation by first computing a sparsely sampled (in scale) pyramid and then interpolating missing levels.
Before HOG and SIFT, early work on face detection with ConvNets \cite{Lecun94,Rowley95} computed shallow networks over image pyramids to detect faces across scales.

\paragraph{Deep ConvNet object detectors.}
With the development of modern deep ConvNets \cite{Krizhevsky2012}, object detectors like OverFeat \cite{Sermanet2014} and R-CNN \cite{Girshick2014} showed dramatic improvements in accuracy.
OverFeat adopted a strategy similar to early neural network face detectors by applying a ConvNet as a sliding window detector on an image pyramid.
R-CNN adopted a region proposal-based strategy \cite{Uijlings2013} in which each proposal was scale-normalized before classifying with a ConvNet.
SPPnet \cite{He2014} demonstrated that such region-based detectors could be applied much more efficiently on feature maps extracted on a single image scale.
Recent and more accurate detection methods like Fast R-CNN \cite{Girshick2015a} and Faster R-CNN \cite{Ren2015a} advocate using features computed from a single scale, because it offers a good trade-off between accuracy and speed.
Multi-scale detection, however, still performs better, especially for small objects.

\paragraph{Methods using multiple layers.}
A number of recent approaches improve detection and segmentation by using different layers in a ConvNet.
FCN \cite{Long2015} sums partial scores for each category over multiple scales to compute semantic segmentations.
Hypercolumns \cite{Hariharan2015} uses a similar method for object instance segmentation.
Several other approaches (HyperNet \cite{Kong2016}, ParseNet \cite{Liu2015}, and ION \cite{Bell2016}) concatenate features of multiple layers before computing predictions, which is equivalent to summing transformed features.
SSD \cite{Liu2016} and MS-CNN \cite{Cai2016} predict objects at multiple layers of the feature hierarchy without combining features or scores.

There are recent methods exploiting lateral/skip connections that associate low-level feature maps across resolutions and semantic levels, including U-Net \cite{ronneberger2015} and SharpMask \cite{Pinheiro2016} for segmentation, Recombinator networks \cite{Honari2016} for face detection, and Stacked Hourglass networks \cite{Newell2016} for keypoint estimation.
Ghiasi \etal \cite{Ghiasi2016} present a Laplacian pyramid presentation for FCNs to progressively refine segmentation.
Although these methods adopt architectures with pyramidal shapes, they are unlike featurized image pyramids \cite{Dalal2005,Felzenszwalb2010,Sermanet2014} where predictions are made independently at all levels, see Fig.~\ref{fig:multilevel}. In fact, for the pyramidal architecture in Fig.~\ref{fig:multilevel} (top), image pyramids are still needed to recognize objects across multiple scales \cite{Pinheiro2016}.

\section{Feature Pyramid Networks}
\label{sec:fpn}

Our goal is to leverage a ConvNet's pyramidal feature hierarchy, which has semantics from low to high levels, and build a feature pyramid with high-level semantics throughout.
The resulting \emph{Feature Pyramid Network} is general-purpose and in this paper we focus on sliding window proposers (Region Proposal Network, RPN for short) \cite{Ren2015a} and region-based detectors (Fast R-CNN) \cite{Girshick2015a}.
We also generalize FPNs to instance segmentation proposals in Sec.~\ref{sec:segment}.

Our method takes a single-scale image of an arbitrary size as input, and outputs proportionally sized feature maps at multiple levels, in a fully convolutional fashion.
This process is independent of the backbone convolutional architectures (\eg, \cite{Krizhevsky2012,Simonyan2015,He2016}), and in this paper we present results using ResNets \cite{He2016}.
The construction of our pyramid involves a bottom-up pathway, a top-down pathway, and lateral connections, as introduced in the following.

\paragraph{Bottom-up pathway.}
The bottom-up pathway is the feedforward computation of the backbone ConvNet, which computes a feature hierarchy consisting of feature maps at several scales with a scaling step of 2.
There are often many layers producing output maps of the same size and we say these layers are in the same network \emph{stage}.
For our feature pyramid, we define one pyramid level for each stage.
We choose the output of the last layer of each stage as our reference set of feature maps, which we will enrich to create our pyramid.
This choice is natural since the deepest layer of each stage should have the strongest features.

Specifically, for ResNets \cite{He2016} we use the feature activations output by each stage's last residual block.
We denote the output of these last residual blocks as $\{C_2, C_3, C_4, C_5\}$ for conv2, conv3, conv4, and conv5 outputs, and note that they have strides of \{4, 8, 16, 32\} pixels with respect to the input image.
We do not include conv1 into the pyramid due to its large memory footprint.

\paragraph{Top-down pathway and lateral connections.}
The top-down pathway hallucinates higher resolution features by upsampling spatially coarser, but semantically stronger, feature maps from higher pyramid levels.
These features are then enhanced with features from the bottom-up pathway via lateral connections.
Each lateral connection merges feature maps of the same spatial size from the bottom-up pathway and the top-down pathway.
The bottom-up feature map is of lower-level semantics, but its activations are more accurately localized as it was subsampled fewer times.

\begin{figure}[t]
\center
\includegraphics[width=0.6\linewidth]{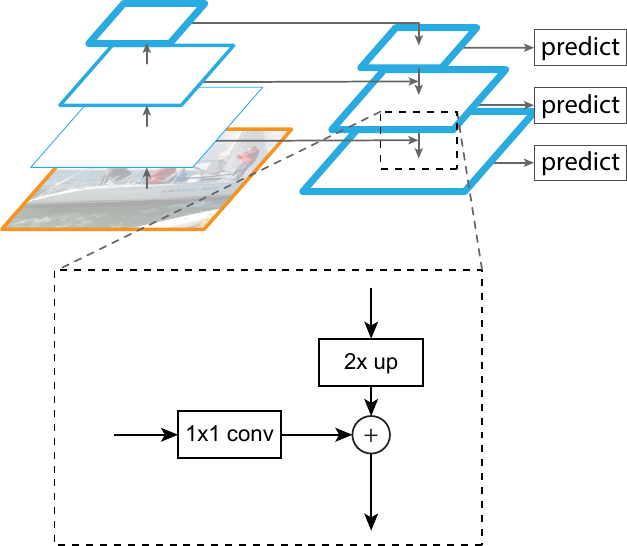}
\caption{A building block illustrating the lateral connection and the top-down pathway, merged by addition.}
\label{fig:block}
\end{figure}

Fig.~\ref{fig:block} shows the building block that constructs our top-down feature maps.
With a coarser-resolution feature map, we upsample the spatial resolution by a factor of 2 (using nearest neighbor upsampling for simplicity).
The upsampled map is then merged with the corresponding bottom-up map (which undergoes a 1$\times$1 convolutional layer to reduce channel dimensions) by element-wise addition.
This process is iterated until the finest resolution map is generated.
To start the iteration, we simply attach a 1$\times$1 convolutional layer on $C_5$ to produce the coarsest resolution map.
Finally, we append a $3\times 3$ convolution on each merged map to generate the final feature map, which is to reduce the aliasing effect of upsampling.
This final set of feature maps is called $\{P_2, P_3, P_4, P_5\}$, corresponding to $\{C_2, C_3, C_4, C_5\}$ that are respectively of the same spatial sizes.

Because all levels of the pyramid use shared classifiers/regressors as in a traditional featurized image pyramid, we fix the feature dimension (numbers of channels, denoted as $d$) in all the feature maps.
We set $d=256$ in this paper and thus all extra convolutional layers have 256-channel outputs.
There are no non-linearities in these extra layers, which we have empirically found to have minor impacts.

Simplicity is central to our design and we have found that our model is robust to many design choices.
We have experimented with more sophisticated blocks (\eg, using multi-layer residual blocks \cite{He2016} as the connections) and observed marginally better results.
Designing better connection modules is not the focus of this paper, so we opt for the simple design described above. 

\section{Applications}

Our method is a generic solution for building feature pyramids inside deep ConvNets.
In the following we adopt our method in RPN \cite{Ren2015a} for bounding box proposal generation and in Fast R-CNN \cite{Girshick2015a} for object detection.
To demonstrate the simplicity and effectiveness of our method, we make minimal modifications to the original systems of \cite{Ren2015a,Girshick2015a} when adapting them to our feature pyramid.

\subsection{Feature Pyramid Networks for RPN}

RPN \cite{Ren2015a} is a sliding-window class-agnostic object detector.
In the original RPN design, a small subnetwork is evaluated on dense 3$\times$3 sliding windows, on top of a single-scale convolutional feature map, performing object/non-object binary classification and bounding box regression.
This is realized by a 3$\times$3 convolutional layer followed by two sibling 1$\times$1 convolutions for classification and regression, which we refer to as a network \emph{head}.
The object/non-object criterion and bounding box regression target are defined with respect to a set of reference boxes called \emph{anchors} \cite{Ren2015a}.
The anchors are of multiple pre-defined scales and aspect ratios in order to cover objects of different shapes.

We adapt RPN by replacing the single-scale feature map with our FPN.
We attach a head of the same design (3$\times$3 conv and two sibling 1$\times$1 convs) to each level on our feature pyramid.
Because the head slides densely over all locations in all pyramid levels, it is not necessary to have multi-scale anchors on a specific level.
Instead, we assign anchors of a single scale to each level.
Formally, we define the anchors to have areas of $\{32^2, 64^2, 128^2, 256^2, 512^2\}$ pixels on $\{P_2, P_3, P_4, P_5, P_6\}$ respectively.\footnote{Here we introduce $P_6$ only for covering a larger anchor scale of $512^2$. 
$P_6$ is simply a stride two subsampling of $P_5$. $P_6$ is not used by the Fast R-CNN detector in the next section.}
As in \cite{Ren2015a} we also use anchors of multiple aspect ratios $\{1$:$2,$ $1$:$1$, $2$:$1\}$ at each level.
So in total there are 15 anchors over the pyramid.

We assign training labels to the anchors based on their Intersection-over-Union (IoU) ratios with ground-truth bounding boxes as in \cite{Ren2015a}.
Formally, an anchor is assigned a positive label if it has the highest IoU for a given ground-truth box or an IoU over 0.7 with any ground-truth box, and a negative label if it has IoU lower than 0.3 for all ground-truth boxes.
Note that scales of ground-truth boxes are not explicitly used to assign them to the levels of the pyramid; instead, ground-truth boxes are associated with anchors, which have been assigned to pyramid levels. As such, we introduce no extra rules in addition to those in~\cite{Ren2015a}.

We note that the parameters of the heads are shared across all feature pyramid levels;
we have also evaluated the alternative without sharing parameters and observed similar accuracy.
The good performance of sharing parameters indicates that all levels of our pyramid share similar semantic levels.
This advantage is analogous to that of using a featurized image pyramid, where a common head classifier can be applied to features computed at any image scale.

With the above adaptations, RPN can be naturally trained and tested with our FPN, in the same fashion as in \cite{Ren2015a}. We elaborate on the implementation details in the experiments.

\subsection{Feature Pyramid Networks for Fast R-CNN}

Fast R-CNN \cite{Girshick2015a} is a region-based object detector in which Region-of-Interest (RoI) pooling is used to extract features.
Fast R-CNN is most commonly performed on a single-scale feature map.
To use it with our FPN, we need to assign RoIs of different scales to the pyramid levels.

We view our feature pyramid as if it were produced from an image pyramid. Thus we can adapt the assignment strategy of region-based detectors \cite{He2014,Girshick2015a} in the case when they are run on image pyramids. 
Formally, we assign an RoI of width $w$ and height $h$ (on the input image to the network) to the level $P_k$ of our feature pyramid by: 
\begin{equation}\label{eq:roi_mapping}
k=\lfloor k_0+\log_2(\sqrt{wh}/224) \rfloor.
\end{equation}
Here 224 is the canonical ImageNet pre-training size, and $k_0$ is the target level on which an RoI with $w\times h=224^2$ should be mapped into.
Analogous to the ResNet-based Faster R-CNN system \cite{He2016} that uses $C_4$ as the single-scale feature map, we set $k_0$ to 4.
Intuitively, Eqn.~(\ref{eq:roi_mapping}) means that if the RoI's scale becomes smaller (say, 1/2 of 224), it should be mapped into a finer-resolution level (say, $k=3$).

We attach predictor heads (in Fast R-CNN the heads are class-specific classifiers and bounding box regressors) to all RoIs of all levels.
Again, the heads all share parameters, regardless of their levels.
In \cite{He2016}, a ResNet's conv5 layers (a 9-layer deep subnetwork) are adopted as the head on top of the conv4 features, but our method has already harnessed conv5 to construct the feature pyramid.
So unlike \cite{He2016}, we simply adopt RoI pooling to extract 7$\times$7 features, and attach two hidden 1,024-d fully-connected (\emph{fc}) layers (each followed by ReLU) before the final classification and bounding box regression layers.
These layers are randomly initialized, as there are no pre-trained \emph{fc} layers available in ResNets.
Note that compared to the standard conv5 head, our 2-\emph{fc} MLP head is lighter weight and faster.

Based on these adaptations, we can train and test Fast R-CNN on top of the feature pyramid.
Implementation details are given in the experimental section.  

\section{Experiments on Object Detection}

We perform experiments on the 80 category COCO detection dataset \cite{Lin2014}.
We train using the union of 80k train images and a 35k subset of val images (\texttt{trainval35k} \cite{Bell2016}), and report ablations on a 5k subset of val images (\texttt{minival}).
We also report final results on the standard test set (\texttt{test-std}) \cite{Lin2014} which has no disclosed labels.

As is common practice \cite{Girshick2014}, all network backbones are pre-trained on the ImageNet1k classification set \cite{Russakovsky2015} and then fine-tuned on the detection dataset.
We use the pre-trained ResNet-50 and ResNet-101 models that are publicly available.\footnote{\fontsize{7pt}{1em}\url{https://github.com/kaiminghe/deep-residual-networks}}
Our code is a reimplementation of \texttt{py-faster-rcnn}\footnote{\fontsize{7pt}{1em}\url{https://github.com/rbgirshick/py-faster-rcnn}} using Caffe2.\footnote{\fontsize{7pt}{1em}\url{https://github.com/caffe2/caffe2}}

\newcommand{\pyr}{$\{P_k\}$}

\newcolumntype{x}[1]{p{#1pt}}
\setlength{\tabcolsep}{4pt}
\renewcommand{\arraystretch}{1.1}
\begin{table*}[t]
\begin{center}
\footnotesize
\begin{tabular}{x{120}|c|c|cc|c|c|ccc}
\textbf{RPN} & feature & \# anchors & lateral? & top-down? & 
 AR$^{100}$ &  AR$^{1k}$ &
 AR$^{1k}_s$ &  AR$^{1k}_m$ &  AR$^{1k}_l$\\
\shline
(a) baseline on conv4 		& $C_4$ 			& 47k 	&  					&  					& 36.1 			& 48.3				& 32.0 			& 58.7 			& 62.2 \\
(b) baseline on conv5 		& $C_5$ 			& 12k 	&  					&  					& 36.3 			& 44.9				& 25.3 			& 55.5 			& 64.2 \\
(c) \bd{FPN} 				& \pyr			 	& 200k & \checkmark 	& \checkmark 				& \bd{44.0} 	& \bd{56.3} 		& \bd{44.9} 	& \bd{63.4} 	& 66.2 \\
\hline
\multicolumn{5}{l}{\emph{\footnotesize Ablation experiments follow:}} \\
\hline
(d) bottom-up pyramid 	& \pyr  				& 200k & \checkmark 	&  					& 37.4 			& 49.5				& 30.5 			& 59.9 			& \textbf{68.0} \\
(e) top-down pyramid, w/o lateral 
										& \pyr			  	& 200k &  					& \checkmark 	& 34.5 			& 46.1 				& 26.5 			& 57.4 			& 64.7 \\
(f) only finest level 			& $P_2$		 		& 750k & \checkmark 	& \checkmark 	& 38.4 			& 51.3 				& 35.1 			& 59.7 			& 67.6 \\
\end{tabular}
\end{center}
\caption{Bounding box proposal results using RPN \cite{Ren2015a}, evaluated on the COCO \texttt{minival} set. All models are trained on \texttt{trainval35k}.
The columns ``lateral'' and ``top-down'' denote the presence of lateral and top-down connections, respectively. The column ``feature'' denotes the feature maps on which the heads are attached. All results are based on ResNet-50 and share the same hyper-parameters.
}
\label{tab:rpn}
\begin{center}
\footnotesize
\begin{tabular}{x{120}|l|c|c|cc|c|c|ccc}
\textbf{Fast R-CNN} & proposals & feature & head & lateral? & top-down? & 
 AP@0.5 &  AP &
 AP$_s$ &  AP$_m$ &  AP$_l$\\
\shline
(a) baseline on conv4 		& RPN, \pyr 		& $C_4$ 			& conv5 			&  					& 						& 54.7 	& 31.9 	& 15.7	& 36.5 	& 45.5 \\
(b) baseline on conv5 		& RPN, \pyr 		& $C_5$ 			& 2\emph{fc} 	& 						& 						& 52.9	& 28.8	& 11.9	& 32.4	& 43.4 \\
(c) \bd{FPN}  				& RPN, \pyr 		& \pyr 	& 2\emph{fc} 	& \checkmark 	& \checkmark 	& \textbf{56.9} & \textbf{33.9} & \textbf{17.8} & \textbf{37.7} & \textbf{45.8} \\
\hline
\multicolumn{5}{l}{\emph{\footnotesize Ablation experiments follow:}} \\
\hline
(d) bottom-up pyramid 	& RPN, \pyr 		& \pyr 	& 2\emph{fc} 	& \checkmark 	&  					& 44.9	& 24.9	& 10.9	& 24.4	& 38.5 \\
(e) top-down pyramid, w/o lateral
										& RPN, \pyr 		& \pyr 	& 2\emph{fc} 	&  					& \checkmark 	& 54.0	& 31.3	& 13.3	& 35.2	& 45.3 \\
(f) only finest level 			& RPN, \pyr 		& $P_{2}$ 		& 2\emph{fc} 	& \checkmark 	& \checkmark 	& 56.3	& 33.4	& 17.3	& 37.3	& 45.6 \\
\end{tabular}
\end{center}
\caption{Object detection results using \textbf{Fast R-CNN} \cite{Girshick2015a} on \emph{a fixed set of proposals} (RPN, \pyr, Table~\ref{tab:rpn}(c)), evaluated on the COCO \texttt{minival} set. Models are trained on the \texttt{trainval35k} set. All results are based on ResNet-50 and share the same hyper-parameters.
}
\label{tab:frcn}
\begin{center}
\footnotesize
\begin{tabular}{x{120}|l|c|c|cc|c|c|ccc}
\textbf{Faster R-CNN} & proposals & feature & head & lateral? & top-down? & 
 AP@0.5 &  AP &
 AP$_s$ &  AP$_m$ &  AP$_l$\\
\shline
(*) baseline from He \etal \cite{He2016}$^\dagger$ & RPN, $C_{4}$ 		& $C_4$ 			& conv5 			&  					& 						& 47.3 	& 26.3 	& -	& -	& - \\ 
\hline
(a) baseline on conv4 		& RPN, $C_{4}$ 		& $C_4$ 			& conv5 			&  					& 						& 53.1 	& 31.6 	& 13.2	& 35.6 	& \textbf{47.1} \\
(b) baseline on conv5 		& RPN, $C_{5}$ 		& $C_5$ 			& 2\emph{fc} 	& 						& 						& 51.7	& 28.0	& 9.6	& 31.9	& 43.1 \\
(c) \bd{FPN}  				& RPN, \pyr 		& \pyr 	& 2\emph{fc} 	& \checkmark 	& \checkmark 	& \textbf{56.9} & \textbf{33.9} & \textbf{17.8} & \textbf{37.7} & 45.8 \\
\end{tabular}
\end{center}
\caption{Object detection results using \textbf{Faster R-CNN} \cite{Ren2015a} evaluated on the COCO \texttt{minival} set. \emph{The backbone network for RPN are consistent with Fast R-CNN.}
Models are trained on the \texttt{trainval35k} set and use ResNet-50.
$^\dagger$Provided by authors of \cite{He2016}.
}
\label{tab:fasterrcnn}
\end{table*}

\subsection{Region Proposal with RPN}
\label{sec:exp_rpn}

We evaluate the COCO-style Average Recall (AR) and AR on small, medium, and large objects (AR$_s$, AR$_m$, and AR$_l$) following the definitions in \cite{Lin2014}.
We report results for 100 and 1000 proposals per images (AR$^{100}$ and AR$^{1k}$).

\paragraph{Implementation details.}
All architectures in Table~\ref{tab:rpn} are trained end-to-end.
The input image is resized such that its shorter side has 800 pixels.
We adopt synchronized SGD training on 8 GPUs.
A mini-batch involves 2 images per GPU and 256 anchors per image.
We use a weight decay of 0.0001 and a momentum of 0.9.
The learning rate is 0.02 for the first 30k mini-batches and 0.002 for the next 10k.
For all RPN experiments (including baselines), we include the anchor boxes that are outside the image for training, which is unlike \cite{Ren2015a} where these anchor boxes are ignored.
Other implementation details are as in \cite{Ren2015a}.
Training RPN with FPN on 8 GPUs takes about 8 hours on COCO.

\subsubsection{Ablation Experiments}

\paragraph{Comparisons with baselines.} For fair comparisons with original RPNs \cite{Ren2015a}, we run two baselines (Table~\ref{tab:rpn}(a, b)) using the single-scale map of $C_4$ (the same as \cite{He2016}) or $C_5$, both using the same hyper-parameters as ours, including using 5 scale anchors of $\{32^2, 64^2, 128^2, 256^2, 512^2\}$.
Table~\ref{tab:rpn} (b) shows no advantage over (a), indicating that a single higher-level feature map is not enough because there is a trade-off between coarser resolutions and stronger semantics.

Placing FPN in RPN improves AR$^{1k}$ to 56.3 (Table~\ref{tab:rpn} (c)), which is \textbf{8.0} points increase over the single-scale RPN baseline (Table~\ref{tab:rpn} (a)).
In addition, the performance on small objects (AR$^{1k}_s$) is boosted by a large margin of 12.9 points.
Our pyramid representation greatly improves RPN's robustness to object scale variation.

\paragraph{How important is top-down enrichment?}
Table~\ref{tab:rpn}(d) shows the results of our feature pyramid without the top-down pathway.
With this modification, the 1$\times$1 lateral connections followed by 3$\times$3 convolutions are attached to the bottom-up pyramid.
This architecture simulates the effect of reusing the pyramidal feature hierarchy (Fig.~\ref{fig:teaser}(b)).

The results in Table~\ref{tab:rpn}(d) are just on par with the RPN baseline and lag far behind ours.
We conjecture that this is because there are large semantic gaps between different levels on the bottom-up pyramid (Fig.~\ref{fig:teaser}(b)), especially for very deep ResNets.
We have also evaluated a variant of Table~\ref{tab:rpn}(d) without sharing the parameters of the heads, but observed similarly degraded performance.
This issue cannot be simply remedied by level-specific heads.

\paragraph{How important are lateral connections?}
Table~\ref{tab:rpn}(e) shows the ablation results of a top-down feature pyramid without the 1$\times$1 lateral connections.
This top-down pyramid has strong semantic features and fine resolutions.
But we argue that the locations of these features are not precise, because these maps have been downsampled and upsampled several times.
More precise locations of features can be directly passed from the finer levels of the bottom-up maps via the lateral connections to the top-down maps.
As a results, FPN has an AR$^{1k}$ score 10 points higher than Table~\ref{tab:rpn}(e).

\paragraph{How important are pyramid representations?}
Instead of resorting to pyramid representations, one can attach the head to the highest-resolution, strongly semantic feature maps of $P_2$ (\ie, the finest level in our pyramids).
Similar to the single-scale baselines, we assign all anchors to the $P_2$ feature map.
This variant (Table~\ref{tab:rpn}(f)) is better than the baseline but inferior to our approach.
RPN is a sliding window detector with a fixed window size, so scanning over pyramid levels can increase its robustness to scale variance.

In addition, we note that using $P_2$ alone leads to more anchors (750k, Table~\ref{tab:rpn}(f)) caused by its large spatial resolution. This result suggests that a larger number of anchors is not sufficient in itself to improve accuracy.

\subsection{Object Detection with Fast/Faster R-CNN}
\label{sec:frcn_details}

Next we investigate FPN for region-based (non-sliding window) detectors. 
We evaluate object detection by the COCO-style Average Precision (AP) and PASCAL-style AP (at a single IoU threshold of 0.5).
We also report COCO AP on objects of small, medium, and large sizes (namely, AP$_s$, AP$_m$, and AP$_l$) following the definitions in \cite{Lin2014}.

\newcolumntype{x}[1]{p{#1pt}}
\setlength{\tabcolsep}{3.5pt}
\renewcommand{\arraystretch}{1.1}
\begin{table*}[t]
\begin{center}
\footnotesize
\begin{tabular}{l|c|c|c|c|c|ccc|c|c|ccc}
 & & &
\fontsize{6pt}{1em}\selectfont image & \multicolumn{5}{c|}{\texttt{test-dev}} & \multicolumn{5}{c}{\texttt{test-std}} \\
\cline{5-14}
\fontsize{6pt}{1em}\selectfont method &
\fontsize{6pt}{1em}\selectfont backbone &
\fontsize{6pt}{1em}\selectfont competition &
\fontsize{6pt}{1em}\selectfont pyramid &
 AP$_{@.5}$ &  AP & AP$_s$ &  AP$_m$ &  AP$_l$ &  AP$_{@.5}$ &  AP & AP$_s$ &  AP$_m$ &  AP$_l$\\
\hline
ours, Faster R-CNN on \textbf{FPN} & ResNet-101 & - & & \bd{59.1} & \bd{36.2} & \bd{18.2} & \bd{39.0} & 48.2 & \bd{58.5} & \bd{35.8} & \bd{17.5} & \bd{38.7} & 47.8 \\
\hline
\multicolumn{5}{l}{\emph{\footnotesize Competition-winning \textbf{single-model} results follow:}} \\
\hline
G-RMI$^\dagger$ &  Inception-ResNet	& 2016 &  & - & 34.7 & - & - & - & - & - & - & - & - \\
AttractioNet$^{\ddagger}$ \cite{Gidaris2016}
& {\fontsize{7pt}{1em}\selectfont VGG16 + Wide ResNet}$^{\mathsection}$
								& 2016 & \checkmark & 53.4 & 35.7 & 15.6 & 38.0 & \bd{52.7} & 52.9 & 35.3 & 14.7 & 37.6 & \bd{51.9} \\
Faster R-CNN +++ \cite{He2016} & ResNet-101 & 2015  & \checkmark & 55.7 & 34.9 & 15.6 & 38.7 & 50.9 & - & - & - & - & - \\
Multipath \cite{Zagoruyko2016} \fontsize{6pt}{1em}\selectfont (on \texttt{minival}) & VGG-16 & 2015 & & 49.6 & 31.5 & - & - & - & - & - & - & - & - \\
ION$^\ddagger$ \cite{Bell2016}  & VGG-16 & 2015 & & 53.4 & 31.2 & 12.8 & 32.9 & 45.2 & 52.9 & 30.7 & 11.8 & 32.8 & 44.8 \\
\end{tabular}
\end{center}\vspace{-1mm}
\caption{Comparisons of \textbf{single-model} results on the COCO detection benchmark. Some results were not available on the \texttt{test-std} set, so we also include the \texttt{test-dev} results (and for Multipath \cite{Zagoruyko2016} on \texttt{minival}).
$^\dagger$: \url{http://image-net.org/challenges/talks/2016/GRMI-COCO-slidedeck.pdf}. $^\ddagger$: \url{http://mscoco.org/dataset/\#detections-leaderboard}. $^\mathsection$: This entry of AttractioNet \cite{Gidaris2016} adopts VGG-16 for proposals and Wide ResNet \cite{zagoruyko2016wide} for object detection, so is not strictly a single-model result.}\vspace{-3mm}
\label{tab:sota}
\end{table*}

\paragraph{Implementation details.}
The input image is resized such that its shorter side has 800 pixels.
Synchronized SGD is used to train the model on 8 GPUs.
Each mini-batch involves 2 image per GPU and 512 RoIs per image.
We use a weight decay of 0.0001 and a momentum of 0.9.
The learning rate is 0.02 for the first 60k mini-batches and 0.002 for the next 20k.
We use 2000 RoIs per image for training and 1000 for testing. Training Fast R-CNN with FPN takes about 10 hours on the COCO dataset.

\subsubsection{Fast R-CNN (on fixed proposals)}

To better investigate FPN's effects on the region-based detector alone, we conduct ablations of Fast R-CNN on \emph{a fixed set of proposals}.
We choose to freeze the proposals as computed by RPN on FPN (Table~\ref{tab:rpn}(c)), because it has good performance on small objects that are to be recognized by the detector.
For simplicity we do not share features between Fast R-CNN and RPN, except when specified.

As a ResNet-based Fast R-CNN baseline, following \cite{He2016}, we adopt RoI pooling with an output size of 14$\times$14 and attach all conv5 layers as the hidden layers of the head.
This gives an AP of 31.9 in Table~\ref{tab:frcn}(a).
Table~\ref{tab:frcn}(b) is a baseline exploiting an MLP head with 2 hidden \emph{fc} layers, similar to the head in our architecture. It gets an AP of 28.8, indicating that the 2-\emph{fc} head does not give us any orthogonal advantage over the baseline in Table~\ref{tab:frcn}(a).

Table~\ref{tab:frcn}(c) shows the results of our FPN in Fast R-CNN.
Comparing with the baseline in Table~\ref{tab:frcn}(a), our method improves AP by 2.0 points and small object AP by 2.1 points.
Comparing with the baseline that also adopts a 2\emph{fc} head (Table~\ref{tab:frcn}(b)), our method improves AP by 5.1 points.\footnote{We expect a stronger architecture of the head \cite{Ren2015} will improve upon our results, which is beyond the focus of this paper.}
These comparisons indicate that our feature pyramid is superior to single-scale features for a \emph{region-based} object detector.

Table~\ref{tab:frcn}(d) and (e) show that removing top-down connections or removing lateral connections leads to inferior results, similar to what we have observed in the above subsection for RPN.
It is noteworthy that removing top-down connections (Table~\ref{tab:frcn}(d)) significantly degrades the accuracy, suggesting that Fast R-CNN suffers from using the low-level features at the high-resolution maps.

In Table~\ref{tab:frcn}(f), we adopt Fast R-CNN on the single finest scale feature map of $P_2$.
Its result (33.4 AP) is marginally worse than that of using all pyramid levels (33.9 AP, Table~\ref{tab:frcn}(c)).
We argue that this is because RoI pooling is a warping-like operation, which is less sensitive to the region's scales. 
Despite the good accuracy of this variant, it is based on the RPN proposals of $\{P_k\}$ and has thus already benefited from the pyramid representation.

\subsubsection{Faster R-CNN (on consistent proposals)}

In the above we used a fixed set of proposals to investigate the detectors. But in a Faster R-CNN system \cite{Ren2015a}, the RPN and Fast R-CNN must use \emph{the same network backbone} in order to make feature sharing possible.
Table~\ref{tab:fasterrcnn} shows the comparisons between our method and two baselines, all using \emph{consistent} backbone architectures for RPN and Fast R-CNN.
Table~\ref{tab:fasterrcnn}(a) shows our reproduction of the baseline Faster R-CNN system as described in \cite{He2016}.
Under controlled settings, our FPN (Table~\ref{tab:fasterrcnn}(c)) is better than this strong baseline by \textbf{2.3} points AP and \textbf{3.8} points AP@0.5.

Note that Table~\ref{tab:fasterrcnn}(a) and (b) are baselines that are much stronger than the baseline provided by He \etal \cite{He2016} in Table~\ref{tab:fasterrcnn}(*).
We find the following implementations contribute to the gap:
(i) We use an image scale of 800 pixels instead of 600 in \cite{Girshick2015a,He2016};
(ii) We train with 512 RoIs per image which accelerate convergence, in contrast to 64 RoIs in \cite{Girshick2015a,He2016};
(iii) We use 5 scale anchors instead of 4 in \cite{He2016} (adding $32^2$);
(iv) At test time we use 1000 proposals per image instead of 300 in \cite{He2016}.
So comparing with He \etal's ResNet-50 Faster R-CNN baseline in Table~\ref{tab:fasterrcnn}(*), our method improves AP by 7.6 points and AP@0.5 by 9.6 points.

\setlength{\tabcolsep}{6pt}
\renewcommand{\arraystretch}{1.1}
\begin{table}[t]
\begin{center}
\footnotesize
\begin{tabular}{c|c|c|c|c}
			& \multicolumn{2}{c|}{ResNet-50}
			& \multicolumn{2}{c}{ResNet-101}\\
\hline
share features? & AP$_{@0.5}$ & AP & AP$_{@0.5}$ & AP \\
\shline
no	& 56.9 			& 33.9		& 58.0		&35.0			\\
yes	& \bd{57.2} 			& \bd{34.3} 		& \bd{58.2}		& \bd{35.2}			\\
\end{tabular}
\end{center}
\caption{More object detection results using Faster R-CNN and our FPNs, evaluated on \texttt{minival}. Sharing features increases train time by 1.5$\times$ (using 4-step training \cite{Ren2015a}), but reduces test time.
}\vspace{-3mm}
\label{tab:more}
\end{table}

\paragraph{Sharing features.}
In the above,  for simplicity we do not share the features between RPN and Fast R-CNN.
In Table~\ref{tab:more}, we evaluate sharing features following the 4-step training described in \cite{Ren2015a}.
Similar to \cite{Ren2015a}, we find that sharing features improves accuracy by a small margin. Feature sharing also reduces the testing time.

\paragraph{Running time.}
With feature sharing, our FPN-based Faster R-CNN system has inference time of 0.148 seconds per image on a single NVIDIA M40 GPU for ResNet-50, and 0.172 seconds for ResNet-101.\footnote{These runtimes are updated from an earlier version of this paper.}
As a comparison, the single-scale ResNet-50 baseline in Table~\ref{tab:fasterrcnn}(a) runs at 0.32 seconds.
Our method introduces small extra cost by the extra layers in the FPN, but has a lighter weight head. 
Overall our system is faster than the ResNet-based Faster R-CNN counterpart.
We believe the efficiency and simplicity of our method will benefit future research and applications.

\subsubsection{Comparing with COCO Competition Winners}

We find that our ResNet-101 model in Table~\ref{tab:more} is not sufficiently trained with the default learning rate schedule.
So we increase the number of mini-batches by 2$\times$ at each learning rate when training the Fast R-CNN step. This increases AP on \texttt{minival} to 35.6, without sharing features.
This model is the one we submitted to the COCO detection leaderboard, shown in Table~\ref{tab:sota}.
We have not evaluated its feature-sharing version due to limited time, which should be slightly better as implied by Table~\ref{tab:more}.

Table~\ref{tab:sota} compares our method with the \emph{single-model} results of the COCO competition winners, including the 2016 winner G-RMI and the 2015 winner Faster R-CNN+++.
\emph{Without adding bells and whistles}, our single-model entry has surpassed these strong, heavily engineered competitors.
On the \texttt{test-dev} set, our method increases over the existing best results by \textbf{0.5} points of AP (36.2 \vs 35.7) and \textbf{3.4} points of AP@0.5 (59.1 \vs 55.7).
It is worth noting that our method does not rely on image pyramids and only uses a single input image scale, but still has outstanding AP on small-scale objects. This could only be achieved by high-resolution image inputs with previous methods.

Moreover, our method does \emph{not} exploit many popular improvements, such as iterative regression \cite{Gidaris2015}, hard negative mining \cite{Shrivastava2016}, context modeling \cite{He2016}, stronger data augmentation \cite{Liu2016}, \etc.
These improvements are complementary to FPNs and should boost accuracy further.

Recently, FPN has enabled new top results in \emph{all} tracks of the COCO competition, including detection, instance segmentation, and keypoint estimation. See \cite{he2017mask} for details.

\section{Extensions: Segmentation Proposals}
\label{sec:segment}

\begin{figure}[t]
\centering
\includegraphics[width=1\linewidth]{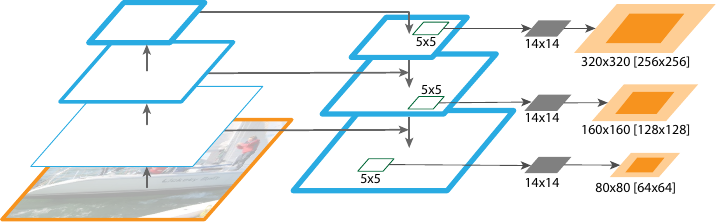}
\caption{FPN for object segment proposals. The feature pyramid is constructed with identical structure as for object detection. We apply a small MLP on 5\x5 windows to generate dense object segments with output dimension of 14\x14. Shown in orange are the size of the image regions the mask corresponds to for each pyramid level (levels $P_{3-5}$ are shown here). Both the corresponding image region size (light orange) and canonical object size (dark orange) are shown. Half octaves are handled by an MLP on 7x7 windows ($7\approx5\sqrt2$), not shown here. Details are in the appendix.
}\vspace{-3mm}
\label{fig:masks}
\end{figure}

Our method is a generic pyramid representation and can be used in applications other than object detection.
In this section we use FPNs to generate segmentation proposals, following the DeepMask/SharpMask framework \cite{Pinheiro2015,Pinheiro2016}.

DeepMask/SharpMask were trained on image crops for predicting instance segments and object/non-object scores.
At inference time, these models are run convolutionally to generate dense proposals in an image.
To generate segments at multiple scales, image pyramids are necessary \cite{Pinheiro2015,Pinheiro2016}.

It is easy to adapt FPN to generate mask proposals.
We use a fully convolutional setup for both training and inference.
We construct our feature pyramid as in Sec.~\ref{sec:exp_rpn} and set $d=128$.
On top of each level of the feature pyramid, we apply a small 5\x5 MLP to predict 14\x14 masks and object scores in a fully convolutional fashion, see Fig.~\ref{fig:masks}.
Additionally, motivated by the use of 2 scales per octave in the image pyramid of \cite{Pinheiro2015,Pinheiro2016}, we use a second MLP of input size 7\x7 to handle half octaves.
The two MLPs play a similar role as anchors in RPN.
The architecture is trained end-to-end; full implementation details are given in the appendix.

\setlength{\tabcolsep}{2pt}
\renewcommand{\arraystretch}{1.1}
\begin{table}[t]
\begin{center}
\footnotesize
\begin{tabular}{l|c|cccc|c}
 & \scriptsize image pyramid & AR & AR$_s$& AR$_m$& AR$_l$ & time (s)  \\
\hline
DeepMask     \cite{Pinheiro2015} & \checkmark & 37.1 & 15.8 & 50.1 & 54.9 & 0.49 \\
SharpMask    \cite{Pinheiro2016} & \checkmark & 39.8 & 17.4 & 53.1 & 59.1 & 0.77 \\
InstanceFCN  \cite{Dai2016a}     & \checkmark & 39.2 & --   & --   & --   & 1.50$^\dagger$ \\
\hline
\multicolumn{7}{l}{\emph{\bd{FPN Mask Results:}}} \\
\hline
single MLP [5\x5]     & & 43.4 & 32.5 & 49.2 & 53.7 & \bd{0.15}\\
single MLP [7\x7]     & & 43.5 & 30.0 & 49.6 & 57.8 & 0.19\\
dual MLP [5\x5, 7\x7]  & & 45.7 & 31.9 & 51.5 & 60.8 & 0.24\\
+ 2x mask resolution & & 46.7 & 31.7 & 53.1 & 63.2 & 0.25\\
+ 2x train schedule  & & \bd{48.1} & \bd{32.6} & \bd{54.2} & \bd{65.6} & 0.25 \\
\end{tabular}
\end{center}
\caption{Instance segmentation proposals evaluated on the first 5k COCO \texttt{val} images. All models are trained on the \texttt{train} set. DeepMask, SharpMask, and FPN use ResNet-50 while InstanceFCN uses VGG-16. DeepMask and SharpMask performance is computed with models available from \url{https://github.com/facebookresearch/deepmask} (both are the `zoom' variants). $^\dagger$Runtimes are measured on an NVIDIA M40 GPU, except the InstanceFCN timing which is based on the slower K40.
}
\label{tab:mask}
\end{table}

\subsection{Segmentation Proposal Results}

Results are shown in Table~\ref{tab:mask}.
We report segment AR and segment AR on small, medium, and large objects, always for 1000 proposals.
Our baseline FPN model with a single 5\x5 MLP achieves an AR of 43.4.
Switching to a slightly larger 7\x7 MLP leaves accuracy largely unchanged.
Using both MLPs together increases accuracy to 45.7 AR.
Increasing mask output size from 14\x14 to 28\x28 increases AR another point (larger sizes begin to degrade accuracy).
Finally, doubling the training iterations increases AR to 48.1.

We also report comparisons to DeepMask~\cite{Pinheiro2015}, SharpMask~\cite{Pinheiro2016}, and InstanceFCN \cite{Dai2016a}, the previous state of the art methods in mask proposal generation.
We outperform the accuracy of these approaches by over \bd{8.3} points AR.
In particular, we nearly double the accuracy on small objects.

Existing mask proposal methods \cite{Pinheiro2015,Pinheiro2016,Dai2016a} are based on densely sampled image pyramids (\eg, scaled by $2^{\{-2:0.5:1\}}$ in \cite{Pinheiro2015,Pinheiro2016}), making them computationally expensive.
Our approach, based on FPNs, is substantially faster (our models run at 6 to 7 FPS).
These results demonstrate that our model is a generic feature extractor and can replace image pyramids for other multi-scale detection problems.

\section{Conclusion}

We have presented a clean and simple framework for building feature pyramids inside ConvNets.
Our method shows significant improvements over several strong baselines and competition winners.
Thus, it provides a practical solution for research and applications of feature pyramids, without the need of computing image pyramids.
Finally, our study suggests that despite the strong representational power of deep ConvNets and their implicit robustness to scale variation, it is still critical to explicitly address multi-scale problems using pyramid representations.

\appendix

\section{Implementation of Segmentation Proposals}

We use our feature pyramid networks to efficiently generate object segment proposals, adopting an image-centric training strategy popular for object detection \cite{Girshick2015a,Ren2015a}. Our FPN mask generation model inherits many of the ideas and motivations from DeepMask/SharpMask \cite{Pinheiro2015,Pinheiro2016}. However, in contrast to these models, which were trained on image crops and used a densely sampled image pyramid for inference, we perform fully-convolutional training for mask prediction on a feature pyramid. While this requires changing many of the specifics, our implementation remains similar in spirit to DeepMask.
Specifically, to define the label of a mask instance at each sliding window, we think of this window as being a crop on the input image, allowing us to inherit definitions of positives/negatives from DeepMask.
We give more details next, see also Fig.~\ref{fig:masks} for a visualization.

We construct the feature pyramid with $P_{2-6}$ using the same architecture as described in Sec.~\ref{sec:exp_rpn}.
We set $d=128$.
Each level of our feature pyramid is used for predicting masks at a different scale.
As in DeepMask, we define the scale of a mask as the max of its width and height.
Masks with scales of $\{32, 64, 128, 256, 512\}$ pixels map to $\{P_2, P_3, P_4, P_5, P_6\}$, respectively, and are handled by a 5\x5 MLP.
As DeepMask uses a pyramid with half octaves, we use a second slightly larger MLP of size 7\x7 ($7\approx5\sqrt2$) to handle half-octaves in our model (\eg, a $128\sqrt2$ scale mask is predicted by the 7\x7 MLP on $P_4$).
Objects at intermediate scales are mapped to the nearest scale in log space.

As the MLP must predict objects at a range of scales for each pyramid level (specifically a half octave range), some padding must be given around the canonical object size.
We use 25\% padding.
This means that the mask output over $\{P_2, P_3, P_4, P_5, P_6\}$ maps to $\{40, 80, 160, 320, 640\}$ sized image regions for the 5\x5 MLP (and to $\sqrt2$ larger corresponding sizes for the 7\x7 MLP).

Each spatial position in the feature map is used to predict a mask at a different location.
Specifically, at scale $P_k$, each spatial position in the feature map is used to predict the mask whose center falls within $2^k$ pixels of that location (corresponding to $\pm1$ cell offset in the feature map).
If no object center falls within this range, the location is considered a negative, and, as in DeepMask, is used only for training the score branch and not the mask branch.

The MLP we use for predicting the mask and score is fairly simple. 
We apply a 5\x5 kernel with 512 outputs, followed by sibling fully connected layers to predict a 14\x14 mask ($14^2$ outputs) and object score (1 output).
The model is implemented in a fully convolutional manner (using 1\x1 convolutions in place of fully connected layers).
The 7\x7 MLP for handling objects at half octave scales is identical to the 5\x5 MLP except for its larger input region.

During training, we randomly sample 2048 examples per mini-batch (128 examples per image from 16 images) with a positive/negative sampling ratio of 1:3.
The mask loss is given 10\x~higher weight than the score loss.
This model is trained end-to-end on 8 GPUs using synchronized SGD (2 images per GPU).
We start with a learning rate of 0.03 and train for 80k mini-batches, dividing the learning rate by 10 after 60k mini-batches.
The image scale is set to 800 pixels during training and testing (we do not use scale jitter).
During inference our fully-convolutional model predicts scores at all positions and scales and masks at the 1000 highest scoring locations.
We do not perform any non-maximum suppression or post-processing.

{\small
\bibliographystyle{ieee}
\bibliography{fpn.bib}
}

\end{document}